\documentclass[
]{ceurart}

\usepackage{listings}
\usepackage{siunitx}  %

\begin{document}

\copyrightyear{2024}
\copyrightclause{Copyright for this paper by its authors.
  Use permitted under Creative Commons License Attribution 4.0
  International (CC BY 4.0).}

\conference{CHR 2024: Computational Humanities Research Conference, December 4–6, 2024, Aarhus, Denmark}

\title{Subversive Characters and Stereotyping Readers}
\subtitle{Characterizing Queer Relationalities with Dialogue-Based Relation Extraction}

\author{Kent K. Chang}[%
orcid=0009-0008-6430-3701,
email=kentkchang@berkeley.edu,
]
\cormark[1]
\address{School of Information, University of California, Berkeley, United States of America}

\author{Anna Ho}[%
email=annaho@berkeley.edu,
]

\author{David Bamman}[%
email=dbamman@berkeley.edu,
]

\cortext[1]{Corresponding author.}

\begin{abstract}
Television is often seen as a site for subcultural identification and subversive fantasy, including in queer cultures.
How might we measure subversion, or the degree to which the depiction of social relationship between a dyad (e.g. two characters who are colleagues) deviates from its typical representation on TV? 
To explore this question, we introduce the task of stereotypic relationship extraction.
Built on cognitive stylistics, linguistic anthropology, and dialogue relation extraction, in this paper, we attempt to model the cognitive process of stereotyping TV characters in dialogic interactions. Given a dyad, we want to predict: what social relationship do the speakers exhibit through their words? Subversion is then characterized by the discrepancy between the distribution of the model's predictions and the ground truth labels. 
To demonstrate the usefulness of this task and gesture at a methodological intervention, we enclose four case studies to characterize the representation of queer relationalities in the \textit{Big Bang Theory}, \textit{Frasier}, and \textit{Gilmore Girls} as we explore the suspicious and reparative modes of reading with our computational methods.
\end{abstract}


\maketitle

\section{Introduction}

Television, often featuring hyper-realized characters, is an important venue for understanding social and relational identities. 
Take this scene from the \textit{Big Bang Theory}:

\begin{quote}
\textsc{howard}. So, who wants to rent \textit{Fiddler}?\\
\textsc{sheldon}. No need! We have the special edition.\\
\textsc{leonard}. Well, maybe we \textit{are} like Haroun and Tanweer. \hfill (season 1, episode 8)
\end{quote}

\noindent Haroun and Tanweer are, as just revealed to the characters, a gay couple who recently adopted a baby.
Knowing that they are a gay couple, Leonard immediately assumes that they love the musical theater that \textit{Fiddler on the Roof} represents.
Indeed, in popular culture, musical theater might have been the \textit{queerest} genre.
However, in his appraisal of its cultural significance in queer culture, John M. Clum, writing in 1999, opens with a rather curious note: ``The surfeit of television situation comedy has pretty much killed stage comedy. When you can get crypto-gay \textit{Frasier} free every week, who needs the gag-rich musical comedy?''~\cite[p. 12]{Clum2001-cc}

Perhaps Clum is right. 
In that same year, his ``crypto-gay'' show aired an episode rather reminiscent of Wildean comedy of manners, in which the Crane brothers try to organize a dinner party (or according to them, an ``intime soiree''), constantly on the phone inviting their friends, only to find this in the voicemail:

\begin{quote}
\textsc{\quad allison}. We just got invited to a dinner party at Dr. Crane's.

\textsc{\quad harry}. Which Dr. Crane?

\textsc{\quad allison}. Does it matter?  You get the one, you get that other one. Personally, I think the whole arrangement's a little $\dots$
\end{quote}

\noindent What Alison truly thinks of them is never revealed to the audience, although the quick-witted Frasier jumps to his conclusion:

\begin{quote}
\textsc{\quad niles}. What you suppose she meant by that?

\textsc{\quad frasier}. She thinks we're always together---that we're some sort of $\dots$ \textit{couple}. 

\textsc{\quad niles}. That's ridiculous! We spend lots of time apart. Besides, who is she to talk? Look at her and Harry! They go everywhere together.

\textsc{\quad frasier}. They're \textit{married}, Niles! Still, there's no reason for her to call us \textit{odd}.

{\raggedleft (season 6, episode 17)\par}

\end{quote}

\noindent What ensues is another similarly frivolous argument on ``who's the other one,'' then another on who not to invite to dinner. 
Much of this episode is Frasier and Niles arguing in the apartment like an old, married couple (``It's possible we have grown a tad dependent on one another.'') while appreciating each other ( ``So we spend a lot of time together---so what? I enjoy it!'').

Those aforementioned scenes from the \textit{Big Bang Theory} and \textit{Frasier} deal with a key aspect of discursive interactions that constitute the subject of the present study: how certain forms of dialogic interaction index certain social and relational identities, including that of a stereotypical gay man or a married couple. 
If Frasier and Niles are \textit{odd} as brothers, who are they to each other?
If we see their speeches as indexical signifiers, what social identities do they anchor in the interactional context?
Built on cognitive stylistics~\cite{Culpeper2001-go} and social semiotics~\cite{Silverstein2022-xc}, 
this work seeks to understand how the depiction of a social relationship between a pair of characters (or a \textit{dyad}) in conversation deviates from the typical representation of that relationship type on TV. 
We hope to advance an operationalization of subversion grounded in queer studies to interrogate the representation of queer relationalities on TV, and enclose four case studies to demonstrate how it can enable more close readings, which has implications for both computational humanities and queer studies.\footnote{See Appendix~\ref{sec:related_work} for more related work.}

\section{Task and data}\label{sec:methods}

\subsection{Task: dialogue-based stereotypic relationship extraction}

Our main computational task involves predicting a stereotypical relationship type, given a dialogue between a dyad (or, two characters) from a scene in a TV series.
We follow convention of relation extraction in NLP~\cite{Yu2020-sv,Jiang2022-bk} and refer to each dyad in terms of \textit{head} and \textit{tail}; unlike the core NLP task, however, we are not simply trying to optimize a model for the true relationship type, but rather identify those moments of dissonance between the truth and a prediction. 
For example, consider this line from \textit{Gilmore Girls}: ``Lorelai, go to your room!'' 
Rory, who says this line, is Lorelai's daughter, but here she sounds like her \textit{mother} for the dramatic effect, and indeed, ``go to your room'' sounds \textit{stereotypically} like a parent.
This kind of subversion is at the core of this work: Rory deviates from the representation of a daughter and talks like a mother.
In terms of modeling, the ground truth relationship type is \texttt{child\_of} for the dyad \texttt{(Rory, Lorelai)}, but we expect the prediction of the stereotypic relationship to be \texttt{parent\_of}, precisely because we are modeling the stereotypic belief~\cite{Culpeper2001-go} of a viewer.
Formally, given a scene $\mathcal{S}$ comprised of multiple dialogue turns, each an utterance $u_i = (s_i, w_0, \dots, w_n)$, where $c \in S_{\mathcal{S}}$ denotes the character (or speaker) identity among all characters $C$ in the scene $\mathcal{S}$, and $w$ the words they speak, we seek to train a model $F(c_h^{\mathcal{S}}$, $c_t^{\mathcal{S}})$, where $c_h$ is the character that occupies the head position, $c_t$ the corresponding tail, to predict a relationship label $r\in\mathcal{R}$, making it an $|\mathcal{R}|$-way classification problem.

\subsection{Dataset: dialogues and dyads}

In order to carry out this inquiry, we need to identify the \textit{true} relationship for a set of character pairs attended with dialogue in scripts, and build a model of their stereotypical interaction.
To support this task, we put together a dedicated dataset of the following components:

\paragraph{Dialogues from parsed teleplays.} We digitize and parse teleplays of pilot episodes from TV Writing to support this task.\footnote{\url{https://sites.google.com/site/tvwriting/}.}
Pilot episodes in the teleplay format are preferable for this task because, unlike sampling a random episode from a collection of transcripts, they typically do not require their audience to have any background knowledge of the series itself, and the standardized format of the teleplay gives us reliable scene segmentation as well as other structural information of the interaction (e.g. speaker labels and background action statements). 
However, digitized PDF files are unstructured. 
To address this, we leverage the fact that in teleplays, structural elements are distinguished by the amount of indentation a line has. For example, speaker labels are most heavily indented.
We used Teseract to perform OCR on the PDF files,\footnote{\url{https://github.com/tesseract-ocr/tesseract}.} which gives us both the recognized texts and the bounding box associated with them for each page.
With this information, we used OpenAI's GPT-4o to parse those OCR'd teleplays in a one-shot setup: the model is tasked to classify each line as one of the following: scene header, speaker label, speaker note, action statement, or other, and combine consecutive lines of the same structural role.
This results in 787 titles, which we split into training, development, and test sets. 

\paragraph{Relationship type labels.} Another component is the relationship type labels that we can use for training models and analysis. 
We first use Wikipedia On-Demand API to identify and gather pages related to each title for which we have the teleplay of the pilot episode.\footnote{\url{https://enterprise.wikimedia.com/products/}.} 
With this resource, we devise a pipeline of three stages:
During the \textit{mining} stage, we employ GPT-4o to extract relationship tuples from these summaries without using any predefined relationship labels. 
This extraction process was followed by a manual review, where we pruned the labels by merging semantically similar ones, such as ``\texttt{kid\_of}'' and ``\texttt{child\_of}''; this is the \textit{pruning} stage. 
To ensure quality, a co-author replicated the relationship extraction task on a test set comprising 50 titles in our test set.
For relations that GPT-4o did extract, the accuracy is satisfactory at $81.67\%$.
Finally, we verify the accuracy of all extractions in both the development and test sets, which concludes the \textit{verification} stage.
In principle, several of the relationship types can be overlapping---two characters may be seen to be both married (\texttt{spouse\_of}) and lovers (\texttt{love\_interest\_of}). 
In order to create a multiclass classification problem (predicting only one label for each dyad), we rank all relationship types based on their specificity (as listed in Table~\ref{tab:relationship_types} in Appendix~\ref{sec:rel_types}) and ask a model to predict the most \emph{specific} relationship type from this set. 
The most frequent relationship types are: \texttt{colleague\_of}, \texttt{friend\_of}, \texttt{sibling\_of}, \texttt{spouse\_of}, and \texttt{classmate\_of}.

\paragraph{Dyads.} To generate dyads for training and analysis, we begin by collecting the set of speaker labels present in each teleplay. 
These labels are often noisy due to OCR errors and inconsistencies with how names appear in Wikipedia summaries. 
To standardize these labels, we query the title in the Movie Database (TMDb),\footnote{\url{https://www.themoviedb.org/}.} which provides a list of characters, both recurring and guest. 
We then match each speaker label with a canonical character name from TMDb using a simple heuristic: if there is at least one token overlap, we select the TMDb character name with minimal edit distance. 
Labels that do not match (typically generic names like ``MAN \#1'') are discarded. 
With standardized speaker labels, we create the list of dyads of interest for each scene by permuting the labels in pairs (i.e. creating 2-permutations from the set of distinct speakers in a scene). 
This means that a pair of characters is considered only if both have speaking roles within the same scene. 
Finally, we include a dyad in our dataset if we have previously extracted its relationship type from Wikipedia. 
See Appendix~\ref{sec:data} for statistics and more information.

\begin{figure}%
  \centering
\begin{lstlisting}[frame=single, basicstyle=\ttfamily\footnotesize, breaklines=true, escapeinside={(*@}{@*)}]
<background> Briefcase in hand , ENTITY 5 is once again waiting for the elevator . He 's approached by ENTITY 6 , 39 , smart , cute , but not sweet . You do n 't get to be hospital administrator and dean of medicine by being sweet .
<speaker> ENTITY 6 <line> I was expecting you in my office twenty minutes ago .
<speaker> ENTITY 5 <line> Really ? That 's odd because I had no intention of being in your office twenty minutes ago .
\end{lstlisting}
  \caption{Example of an anonymized and post-processed scene (only first two lines  represented here).}\label{fig:house_tagged}
\end{figure}

\paragraph{Anonymization.} Previous work indicates speaker anonymization is beneficial as it mitigates the distraction and noise named entities might introduce~\cite{Chen2022-fg}.
We similarly anonymize our dataset to evaluate its impact. 
For each scene, we maintain a mapping table between canonical speaker names and randomly assigned entity IDs. 
All speaker identities are then replaced with \texttt{ENTITY} plus their unique scene ID, and their mentions in dialogue lines are anonymized in the same way.
Some canonical names include generic words, often related to a profession  (e.g. ``coach''), %
and those words would be anonymized, which might lead to an unnecessary loss of information.
To address this, we aggregate all tokens in canonical speaker names and manually annotate whether each token is part of a proper name or a non-name content word. 
An example is in Fig.~\ref{fig:house_tagged} (special tokens like \texttt{<speaker>} are explained in Sec.~\ref{sec:exp}).

\section{Models and experiments}\label{sec:exp}

Given this data, we can build and evaluate models of \emph{stereotyping readers}, who learn from the totality of relationships encoded in our training data to infer the relationship enacted between two characters in a specific scene.  We do not expect a model to be able to identify the \textit{true} relationship with perfect accuracy---not every relationship is enacted in dialogue at the level of a single scene; but more importantly, our work is premised on the idea that the relationship we observe on screen (and that a model observes as well) can exhibit variation that is deliberately at odds with that ``truth.''  And yet, accuracy at predicting that truth provides us with an instrumental means to select between different models, and we assess several variants in order to find the one most sensitive to the dialogic indicators of how relationships are performed.

\subsection{Models}

We compare the performance of the following models on this task to select the best strategy to model stereotyping readers with our data.
We first establish the \textbf{majority class} (predicting the most frequent class, \texttt{colleague\_of}) baseline for all models.
Since the task takes the form $F(c_h^{\mathcal{S}}, c_t^{\mathcal{S}})$, those models can be categorized based on how we choose to represent the character $c_h$ and $c_t$ in scene $\mathcal{S}$:
\textit{supervised} and \textit{prompting}:

\paragraph{Supervised.} We start with an \textit{utterance-based} model, where we string together all utterances spoken by the head and tail speakers, respectively. 
The motivation is that each speaker can be represented by all of their utterances in the scene.
Those utterances are subsequently encoded by the same Longformer--base encoder~\cite{Beltagy2020-vx}; we extract the CLS tokens (\texttt{<s>}) that represent all the utterances and concatenate them, resulting in the overall representation:
\begin{equation}
	\boldsymbol{h} = [\boldsymbol{e}_{\text{\texttt{<s>}}}^{c_h}; \boldsymbol{e}_{\text{\texttt{<s>}}}^{c_t}].
\end{equation}
Next, we include a representation of the entire scene using the \textit{attentive pooling} technique. 
Here, we similarly string all the tokens in the entire scene together, but to enhance the structural awareness of the teleplay (e.g. some tokens are speaker labels, and some are their lines), we introduce the following special tokens, whose representations are learned during training: \texttt{<scene>}, \texttt{<speaker>}, \texttt{<line>}, and \texttt{<background>}.
A full example is in Fig.~\ref{fig:house_tagged}.
We take inspiration from~\cite{Sang2022-fv} and incorporate attentive pooling techniques for the scene representation.
Since we have the utterances from both head and tail speakers from the utterance-based model, we want to emphasize other information in the scene by guiding the model to attend less to those utterances and more to dialogue lines from other speakers.
In encoding the scene here, we introduce a token-level mask $M$, where $M[j] = 0$ if the $j$-th word is spoken by either head or tail speaker and $M[j] = 1$ otherwise. 
Following~\cite{Sang2022-fv}, the scene information selected by $M$ is:
\begin{align}
	\boldsymbol{h}_{\mathcal{S}} = \boldsymbol{e}_{\text{\texttt{<s>}}}^{\mathcal{S}};\qquad 
	A = {\boldsymbol{w}_{A}}^\top \boldsymbol{h}_{\mathcal{S}}; \qquad 
	\alpha = \mathrm{softmax}(A\odot M).
\end{align}
\noindent The head- and tail-aware attention is used to pull the hidden states: ${\boldsymbol{h}_{\mathcal{S}}}^\top \alpha$, which is concatenated with head and tail utterances:
\begin{equation}
		\boldsymbol{h} = [\boldsymbol{e}_{\text{\texttt{<s>}}}^{c_h}; \boldsymbol{e}_{\text{\texttt{<s>}}}^{c_t}; {\boldsymbol{h}_{\mathcal{S}}}^\top \alpha].
\end{equation}
The overall representation $\boldsymbol{h}$ is then fed to a linear classification head $f$, which yields $P(r|\boldsymbol{h}) = \mathrm{softmax}\big( f(\boldsymbol{h}) \big)$ %
and $\hat{r} = \mathrm{arg}\max P(\cdot)$.

\paragraph{Prompting.} For prompt-based models, the overall prompt design resembles a QA task, where given a scene, the model is tasked to answer, \textit{head speaker is \_\_\_\_ of tail speaker}.  
We consider three popular strategies to enhance the performance of large language models: we prompt LLaMA 3--70b--instruct zero-shot and one-shot,\footnote{\url{https://huggingface.co/meta-llama/Meta-Llama-3-70B}.} and leverage the hidden chain-of-thought process~\cite{Wei2022-io} in OpenAI's o1-mini.\footnote{\url{https://openai.com/index/openai-o1-mini-advancing-cost-efficient-reasoning/}.}
For more details, see Appendix~\ref{sec:llamaprompt}.

\subsection{Experimental results}

\begin{table}%
    \centering
    \caption{\label{tab:experiments}
      Experimental results. All metrics are reported with 95\% bootstrap confidence intervals.
    }    
    \setlength{\tabcolsep}{4pt}
    \aboverulesep=0ex %
    \belowrulesep=0ex %
\begin{tabular}{@{}lll}
    \toprule 
                         & \multicolumn{2}{c}{\textbf{Accuracy}}                \\
    \cmidrule(ll){2-3}
                     & anonymized test set & unanonymized test set \\
    \midrule 
    Majority             & 0.265               & 0.265  \\    
    \midrule     
    \multicolumn{3}{@{}l}{\textsc{supervised}} \\[2pt]
    {Longformer}                    & 0.305 {\scriptsize [0.294--0.315]} & 0.298 {\scriptsize [0.287--0.309]} \\
    \;\;{\small + anonymized training set}          & 0.293 {\scriptsize [0.283--0.303]} & 0.306 {\scriptsize [0.295--0.317]}\\
    \;\;{\small + scene attentive pooling} & 0.248 {\scriptsize [0.238--0.258]} & 0.348 {\scriptsize [0.337--0.359]} \\
    \;\;\;\; {\small + both}               & \textbf{0.338} {\scriptsize [0.327--0.349]} & \textbf{0.367} {\scriptsize [0.356--0.378]} \\
    \midrule 
    \multicolumn{3}{@{}l}{\textsc{prompting}} \\[2pt]
    {LLaMA 3--70b}                 & 0.197 {\scriptsize [0.188--0.206]} & 0.243 {\scriptsize [0.233--0.253]} \\
    \;\;{\small + one-shot}               & 0.212 {\scriptsize [0.202--0.221]} & 0.242 {\scriptsize [0.232--0.252]} \\
    {OpenAI o1-mini}               & 0.181 {\scriptsize [0.172--0.190]} & 0.241 {\scriptsize [0.231--0251]} \\
    \bottomrule
\end{tabular}
\end{table}

Experimental results are reported in Table~\ref{tab:experiments}, including 95\% confidence intervals from 10,000 bootstrap resamples.\footnote{For supervised models, we use a learning rate of $5\times 10^{-5}$ with $100$ warm-up steps and no weight decay. 
All inputs are padded or truncated to $4,096$ tokens.
All models are trained on four L40S GPUs.
If the scene is longer than $4,096$ tokens, it is truncated before being inserted into the prompt (only four scenes were truncated).
We use Outline to constrain the output space to be one of the relationship types: \url{https://github.com/outlines-dev/outlines}.} 
We evaluate the performance of each model by comparing its prediction against the true relationship label we have obtained from Wikipedia.
Given the premise of this work is that the true relationship type is not necessarily performed in every interaction, we expect the accuracy here to be low, but the model should still perform better than guessing the most frequent label in the training set.
In assessing the impact of anonymization, we include additional rows for when we anonymize the training set and columns for test ones.

For the supervised models, we observe that adding contextual information about the scene that is absent from the head and tail utterances significantly improves the performance of the model. 
When the model already has the scene information, anonymizing the training set appears beneficial, which is aligned with the findings reported in~\cite{Chen2022-fg}.
For evaluation, anonymization does not significantly hurt the performance of our supervised models.
However, it does for the prompt-based models we evaluate.
While they yield similar performance on the unanonymized data (around $0.242$), the three prompt-based models do not produce identical predictions: LLaMA's zero-shot and one-shot models have a Cohen's $\kappa$~\cite{Cohen1960-sb} of $0.71 [0.70, 0.72]$, and that between LLaMA zero-shot and OpenAI o1-mini is $0.37 [0.36, 0.39]$.
This suggests a low agreement rate between LLaMA and OpenAI models.%

\begin{table}%
    \caption{Most distinct words measured by log-odds ratio for key relationship types.}
    \label{tab:distinct_words}
    \centering
    \begin{tabular}{lp{0.75\linewidth}}%
        \toprule
        \textbf{Relation type} & \textbf{Most distinct words} \\
        \midrule
        \texttt{parent\_of} & kids, son, mother, house, father, darling, honey, debate, worried, stay \\
        \texttt{sibling\_of} & sister, brother, whistledown, lady, cherry, lord, mom, dollars, hastings, must \\
        \texttt{spouse\_of} & honey, love, kids, marriage, baby, married, clean, treat, maple, care \\
        \texttt{colleague\_of} & death, find, magic, heroin, found, real, ship, missing, library, case \\
        \texttt{friend\_of} & girls, fun, york, president, buddy, school, high, rally, jacket, vote \\
        \bottomrule
    \end{tabular}
\end{table}

Since the goal of this section is to figure out the best strategy for modeling stereotyping readers, it is crucial to establish the facial validity of the best-performing model before we impart any trust in its predictions.
In further examining the face validity of the predictions of our best-performing model before moving on to the analysis, we represent the most distinct tokens for the key \emph{predicted} relationship types measured by log-odds ratio with an uninformative Dirichlet prior~\cite{Monroe2017-us} in Table~\ref{tab:distinct_words}, which we can consider a form of post-hoc global explanation providing insight into what a model has learned\ \citep{Du2019-ac}.
We aggregate all head utterances by the {predicted} relationship type and use them as the target corpus, and the rest as the reference corpus.
For each relationship type presented, we include the top ten tokens most strongly associated with the target corpus. 
Those five types are chosen because they are central to our case studies presented in the analysis section below.
In Table~\ref{tab:distinct_words}, we see that the model has learned to associate the relationship types with some of the words the dyad in question typically talks about (e.g. parents talk about kids), \textit{colleague} has to do with occupational terms on TV (e.g. detectives investigate the death of someone), and \textit{friend} is focused on high school life (e.g. students vote for student council president). 
Although there are terms that seem confusing (say, \textit{jacket} in \texttt{colleague\_of}) out of context, the words that are more strongly associated with those categories make an intuitive sense in the context of scripted TV series.

\section{Analysis}

\begin{flushright}
    \begin{minipage}{0.8\textwidth} 
    \small 
        \begin{flushleft}
            Well, I suppose I do think of you as a sister. And sometimes, a mother.
            
            \hspace*{\fill}{---Sheldon Cooper to his friend, Penny, in the \textit{Big Bang Theory}}
        \end{flushleft}

        \begin{flushleft}
            [O]ne of the things that ``queer'' can refer to: the open mesh of possibilities, gaps, overlaps, dissonances and resonances, lapses and excesses of meaning [ . . .].

             \hspace*{\fill}{---\textsc{eve kosofsky sedgwick}, \textit{Tendencies}~\cite{Sedgwick1993-bf}}
        \end{flushleft}
        
    \end{minipage}
\end{flushright}

\noindent Television is often seen as a medium for subcultural identification~\cite{Halperin2012-sa} and subversive fantasy~\cite{Pugh2018-mo}, and this work is animated by the sociological impulse in certain strands of queer studies that see artistic forms as reflecting ``the texture and makeup of queer social worlds''~\cite[p. 115]{Love2021-jd}.
From this perspective, we can see certain moments on TV as representing a queer time and place that sits ``in opposition to the social institutions of family, heterosexuality, and reproduction''~\cite[p. 1]{Jack_Halberstam2005-jl}, where endless possibilities of subversive relational forms take shape in the ``excesses of meaning''~\cite[p. 8]{Sedgwick1993-bf}.
We can use the model described in this paper to characterize this phenomenon. 
For our analysis, we choose the dataset introduced in~\citet{Sang2022-fv}, which consists of five TV series, each almost in their entirety, which allows us to study narrative arcs that span multiple seasons.
Through the case studies, we hope to demonstrate how this work might shed light on the representation of queer modes of relating in the \textit{Big Bang Theory}, \textit{Frasier}, and \textit{Gilmore Girls}.

\subsection{Queer characters and suspicious reading}

Our first case study takes up David Halperin's inquiry into~\textit{queer love}~\cite{Halperin2019-hn} and its central question: ``how is it possible for two men to be together'' when existing social institutions cannot accommodate such a form of togetherness~\citep[p. 136]{Foucault1998-um}.
For Michel Foucault and Halperin, the love between men is queer not because of their sexual preferences and practices; it is instead because of their counter-conduct~\citep{Davidson2011-cw}.
In the context of this study, for any given discursive and dyadic interaction on television, we see how ``one conducts oneself, lets oneself be conducted, and finally, in which one behaves under the influence of a conduct as the action of conducting''~\citep[p. 128]{Davidson2011-cw}.
Conduct is dictated by the social relationship the dyad indexically presumes and projects through conducting with their speech acts.
In this light, the act of subversion through speech---when characters disrupt expected linguistic norms by using forms typically reserved for certain social categories---transgresses the presumptive bounds of relational form and is essential to the practice of counter-conduct.
Importantly, counter-conduct necessitates new forms of relationality: men in queer love need to ``invent, from A to Z, a relationship that is still formless''~\citep[p. 136]{Foucault1998-um}.
Can we use our model to interrogate counter-conduct and relational forms?
The titles in our analysis dataset feature some characters that have become the subject of various queer readings: 
for example, Frasier and Niles from \textit{Frasier} are said to embody gay sensibility~\cite{Raymond2003-yi}, and the four male protagonists of the \textit{Big Bang Theory} perform ``queer-straight masculinity''~\cite{McClanahan2015-mj}.
Juxtaposing the relational possibilities queerness affords and the subversive potentials that many find inherent in those characters, we can use our model as a tool to facilitate and augment close reading.

In the \textit{Big Bang Theory}, Raj Koothrappali and Howard Wolowitz are colleagues at Caltech, but they often talk like a couple: 

\begin{quote}
\textsc{howard}. We're just saying all the things we love about each other.\\
\textsc{raj}. Oh, like you and I did at couple's therapy?\hfill (season 8, episode 9)
\end{quote}

\noindent Much debate surrounding Raj focuses on whether he is gay, to which Steve Molaro, producer and writer of the show, says: while it's viable, ``it was a little more interesting to have a guy so comfortable in his feminine side who’s not gay, and explore that''~\cite[p. 244]{Radloff2022-cm}. 
Along with his self-identification as a metrosexual and fascination with divas (``Cher, Madonna, Adele. All the women who rock me'', season 5 episode 14), we might understand Molaro as gesturing at separating two distinct dimensions of the Raj character: his sexual orientation and, per Foucault, his ``way of life''~\cite{Foucault1998-um}.
Against this backdrop, we argue that Raj and Howard offer more than occasional punchlines; they represent an instance of queer love:
they have to invent a form of togetherness for themselves, despite being constantly under the watch of, occasionally derided by, the maternal signifiers, among them the figure sans figure Mrs. Wolowitz, Howard's mother, who we never see on screen but know she's there: ``After all your sleepovers with the little brown boy, a girl is a big relief'' (season 5, episode 3).
This need to \textit{invent} is made explicit by another mother figure, Dr. Beverly Hofstadter, albeit in a pejorative tone: ``the two of you have created an ersatz homosexual marriage to satisfy your need for intimacy'' (season 2, episode 15).

\begin{figure}%
  \centering
  \includegraphics[width=0.92\linewidth]{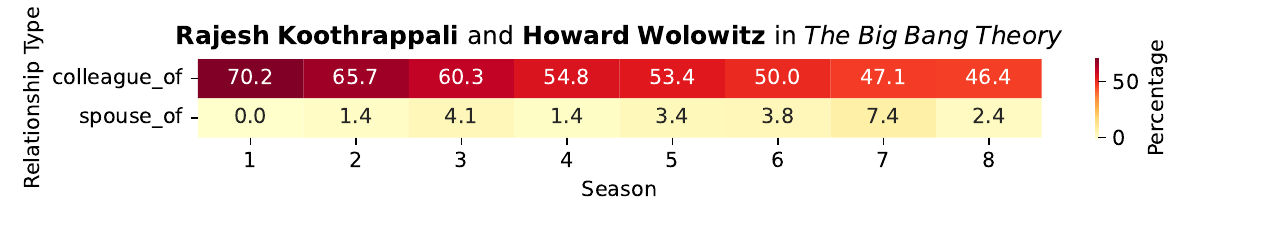} %
  \caption{A heatmap representing the progression of relationship types, ``\texttt{colleague\_of}'' and ``\texttt{spouse\_of}'',  between Raj and Howard in \textit{The Big Bang Theory} across the first eight seasons.}\label{fig:2-1}
\end{figure}

In Fig.~\ref{fig:2-1}, we chart the percentage of the dyad being predicted as \textit{colleagues} or \textit{spouses} to each other across the eight seasons in our dataset of choice. 
It is not surprising to see \textit{colleague} being the most prominent relationship type, but it gradually decreases from $70.2$\% in season 1 to $46.4$\% in Season 8, while the ``\texttt{spouse\_of}" relationship type is predicted starting from season 2, peaking at $7.4$\% in Season 7. 
This indicates a predominant colleague relationship with minor fluctuations towards a more intimate or spousal-like dynamic.
While talking like spouses when they are colleagues can be seen as an act of transgression, do those characters necessarily submit to the social institution of marriage in their form of co-existence, which the \textit{spouse} category would entail?
If we believe our stereotyping reader and assume that the spouse is an observable relational form between them, we ask: can there be subversion within subversion, and do Raj and Howard contest this normative relational form?

For that we return to close, suspcious reading.
One relevant episode takes place towards the end the series (season 12, episode 22): 
We find Raj at an airport, waiting for his flight to London, where he is expected to meet with his girlfriend, Anu.
And we see Howard in discussion on this with his wife, Bernadette:

\begin{quote}
\textsc{bernadette}. Go stop him. Get your best friend back.\\
\textsc{howard}. You \textit{are} my best friend!\\
\textsc{bernadette}. We don't have time for this! Go!
\end{quote}

\noindent This exchange, along with the ensuing airport scene, stages a jubilant celebration of queer love.
If, according to Halperin, ``where the happy couple advances, deviance retreats''~\cite[p. 397]{Halperin2019-hn}, thanks to Bernadette, we see this playing out in the opposite direction: the married couple steps aside, for queer love to flourish.
Howard getting Raj back by no means signals the end of his marriage with Bernadette, but it surfaces the tension between traditional heteronormative relationships and non-normative desires:
Howard's dynamic with Raj reveals an undercurrent of emotional intimacy and dependency that challenges the rigid boundaries of what is considered acceptable, \textit{normal} male bonding within the confines of marriage.
Bernadette, curiously, becomes a kind of referee, simultaneously reinforcing and undermining the normalcy of her marriage: 
She exposes the instability of the heteronormative relationship model, while also ensuring that it doesn't just break apart. 
As Oscar Wilde puts it in his putatively queer \textit{Earnest}, ``In married life, three is company and two is none.''

\begin{figure}%
  \centering
  \includegraphics[width=1.05\linewidth]{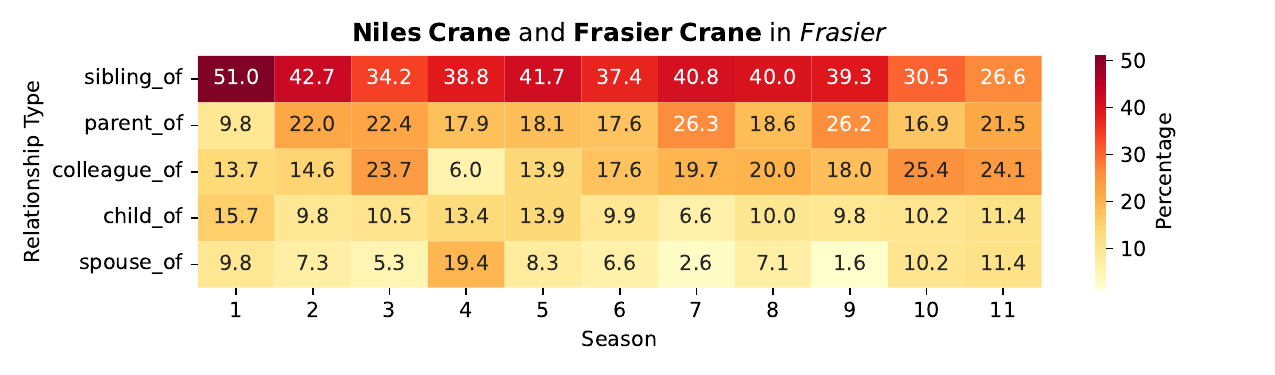} %
  \caption{Another heatmap representing the progression of relationship types between Niles and Frasier in \textit{Frasier} over eleven seasons. Only the top five relationship types are presented here.}\label{fig:2-2}
\end{figure}

Now we return to Niles and Frasier Crane, to the question we raise at the beginning of this paper:
If they are not \textit{just} brothers, who are they to each other?
It is not surprising that in Fig.~\ref{fig:2-2}, the most salient relationship type for them is that of a sibling, but we see how they perform a few different ones over the course of the show.
Towards the end, Nile appears to be just as much a parent and a colleague as a brother for Frasier.
We see how they ``invent'' (in Foucault's word) a new relational form for themselves as they oscillate between the relationship types we already have names for, although others, like Alison and Harry from Introduction, might find them odd---and, indeed, queer---at times. 

\subsection{Anti-normative characters and reparative reading}

Today, \textit{queer} encompasses a broad spectrum of identities, practices, and desires related to sex and gender.
But this was not the case for queer theory at its infancy:
writing around 1993, Eve Kosofsky Sedgwick notes how queer scholarship ``can’t be subsumed under gender and sexuality at all''~\cite[p. 8]{Sedgwick1993-bf}.
In this section, we operate on this expanded notion of queer and turn to subversive modes of parenting.

\begin{figure}%
  \centering
  \includegraphics[width=0.8\linewidth]{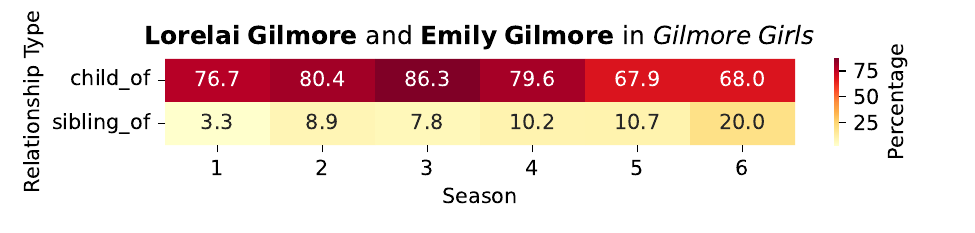} 
  \includegraphics[width=0.8\linewidth]{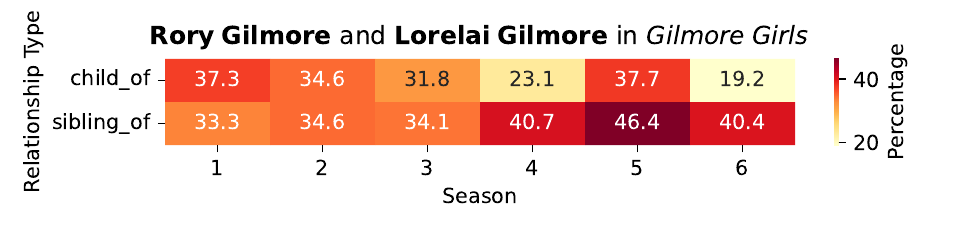}   
  \caption{Heatmap representing the mother--daughter relationship arc between Rory and Lorelai Gilmore (above) and between Lorelai and Emily Gilmore (below) in \textit{Gilmore Girls} in the first six seasons.}\label{fig:1}
\end{figure}

Recall this line from Section~\ref{sec:methods}: ``Go to your room, Lorelai'' (season 1, episode 4) is the daughter talking like a mother.
This is an instance of subversion where traditional parent--child boundaries is blurred, as Rory no longer conforms to the conventional roles. 
We do see two modes of parenting depicted in \textit{Gilmore Girls}: the traditional model Lorelai and her mother, Emily Gilmore represents, and the new, subversive one between Rory and Lorelai.
In Fig.~\ref{fig:1}, we chart the interaction patterns by the percentage of the predicted relationship type between these two pairs of mother and daughter each season, focused on the top three relationship types.
According to our model, we see Emily, compared to Lorelai, is more of a traditional mother throughout, with \texttt{parent\_of} being the dominant relationship type. 
This is aligned with our impression with the characters: as Lorelai says, ``Rory and I are best friends, Mom. We're best friends first, and mother and daughter second. And you and I are mother and daughter always'' (season 2, episode 16).
However, as we see in Fig.~\ref{fig:1}, they start to talk more like sisters around each other in later seasons.
One episode where the dynamic between Lorelai and Emily changes, as identified by our model is ``Friday and Alright For Fighting'' (season 6, episode 13), where Emily says jokingly to Lorelai, ``I only wished I'd remembered to call her a cocktail waitress!'' 
This is a surprise to Lorelai: ``That's my mother's version of the \textit{c} word!''

Another intriguing example of subversive parenting can be found between Sheldon Cooper and Penny in the \textit{Big Bang Theory}.
In this show, Sheldon works with Raj and Howard at Caltech, and Penny represents the ``cute girl next door next to the nerds''~\cite{Dumaraog2021-dk} archetype: the main male cast routinely succumbs to her feminine charm, especially in early seasons, with the sole exception of Sheldon.
For this reason, Sheldon is regarded as asexual~\cite{McClanahan2015-mj} and infantilized~\cite{Shaw2015-mj}.
The latter dimension of the character is explored throughout the series through Penny, which is the main plot of ``The Intimacy Acceleration'' (season 8, episode 15), from which the epigraph of this section is taken.
In this episode, the two engage in a farcical experiment to find out if Sheldon and Penny can fall in love.
For Sheldon, his announced goal is not to win the girl, but to have her drive him to a ``convention celebrating the life and work of Gary Gygax.''
Their experiment commences with Penny saying, ``I will buy you all the dragon T-shirts you want.''
For most of this episode, like a Kleinian baby~\cite{Klein1975-jy}, Sheldon explores object relations and symbol formation in his phantasy-as-experiment: at the end of it, he compares himself to ``human bowl of tomato soup'', and Penny to his sister and his mother.

\begin{figure}%
  \centering
  \includegraphics[width=0.95\linewidth]{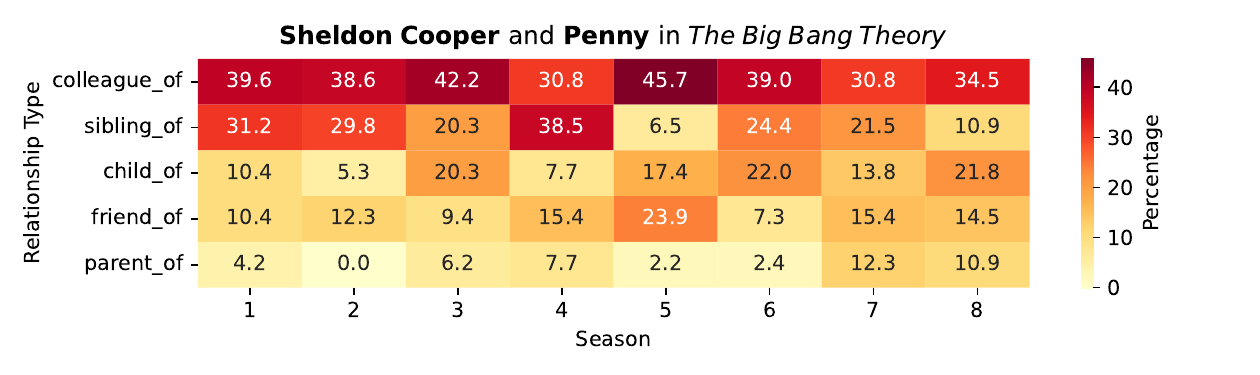} %
  \caption{Another heatmap representing the progression of relationship types between Sheldon Cooper and Penny over eight seasons in the \textit{Big Bang Theory}. Only the top five relationship types are presented here.}\label{fig:sheldon}
\end{figure}

Our model of stereotyping readers captures that, and more. 
The top relationship types for Sheldon and Penny do include \texttt{sibling\_of} and \texttt{child\_of}, and according to the model, one of the moments where Sheldon talks like Penny's child involves her trying to put him to sleep, though unsuccessfully:
``No, I don't want to go to sleep, you can't make me'' (season 8, episode 13).
What's surprising here is Sheldon and Penny, much like Frasier and Niles, take up multiple relational roles over the eight seasons, and the most frequent of them is that of \textit{colleagues}.
If we revisit the series with a keen eye on when they speak like colleagues, we see how Penny can act like a counsel in Sheldon's social and everyday life: on how to empathize with others (season 2, episode 3), girl trouble at a bar where Penny works (season 4, episode 17), and so on.
Penny's capacity for social mentoring reverses the normal dynamics between them, where Sheldon feels and acts like her superior.
In terms of intertextuality, this can explain why Sheldon thinks of Penny as a sister: in the prequel to the series, \textit{Young Sheldon}, we see his sister, Missy Cooper, possess extraordinary social intelligence.

\textit{Surprise} is the operative word for this section, and this choice of words is intentional. 
Inspired by Melanie Klein's work, including her theories on object relations and phantasy, Sedgwick developed her concept of repatative reading: ``[T]o read from a reparative position is to surrender the knowing, anxious paranoid determination that no horror, however apparently unthinkable, shall ever come to the reader as \textit{new}; to a reparatively positioned reader, it can seem realistic and necessary to experience surprise''~\cite[p. 146]{Sedgwick2003-zk}.
In being surprised, we see how computational methods can enable us to take the reparative position as we approach the text; we attempt to bring out the multiplicity of characters, like Emily and Penny, and begin to repair their agency.
Indeed, they are much more than stereotypical mothers or stereotypical cute girls.

\section{Conclusion}

\begin{flushright}
    \begin{minipage}{0.8\textwidth} 
    \small 
        \begin{flushleft}
            A conversation has no necessary terminus.
            
            \hspace*{\fill}{---\textsc{teagan bradway}, ``Queer Narrative Theory and the Relationality of Form''~\cite{Bradway2021-he}}
        \end{flushleft}
    \end{minipage}
\end{flushright}

\noindent In this paper, we model a stereotyping reader who can infer the social relationship of pairs of characters (or dyads) in conversation. 
As an algorithmic measuring device, this model is queer in and of itself:
the model tries to learn, for example, what a couple talks like, but we are ultimately interested in finding characters that are \textit{not} a couple but talk like one, therein lies the queerness we wish to explore.
As such, metrics like accuracy (as we argue in Section~\ref{sec:exp}) are at best an instrumental means, and the model is interesting (as we see in the case studies) only when it predicts anything \textit{but} the truth.
As queer studies intersect with computational humanities, through our model, we wish to take initial steps to ``dismantle the logics of success and failure''~\cite[p. 2]{Halberstam2011-lf} to resist the ``normal business in the academy''~\cite[p. xxvi]{Warner1993-qw}, and in so doing, reflect on disciplinary practices in the relevant research communities to ``build a better description''~\cite{Marcus2016-hc} of queer culture.

One of the powerful images that Sedgwick invokes in her work, which has ultimately altered the landscape of queer theory permeably and forever, is that of a ``theory kindergarten''~\cite[p. 94]{Sedgwick2003-zk}.\footnote{See also~\cite{Britzman2013-wa},~\cite{Love2010-br}.}
As we gesture towards a potential future for a queer cultural analytics~\cite{Chang2023-ff}, we look back on Sedgwick's theory kindergarten. 
Sedgwick was critical, but we might think of a theory kindergarten as a generative space of playful exploration, not bound by rigid epistemological frameworks or normative, institutional constraints.
As such, theory kindergarten can be inclusive, reflexive, and enabling: as we see in \textit{Touching Feeling}, it invites us to ``think otherwise''~\cite[p. 11]{Sedgwick2003-zk}: embrace surprise and creativity, allow for new modes of understanding and inquiry to emerge---perhaps including computation, without reducing the complexity and multiplicity of queer thought.
If relational forms that emerge out of dialogic interactions showcase the ``open mesh of possibilities'' of queerness~\cite[p. 7]{Sedgwick1993-bf} that Sedgwick speaks of, interdisciplinary conversations, as Bradway, the source of the epigraph for this section, so eloquently intimates in the context of queer narrative theory, do not need an overdetermined terminus.
We hope this work can motivate more researchers to study the representation of subversion in queer cultures and test the limits of both queer theory and computational methods.
Back in the kindergarten: experiment and play, think and operationalize otherwise, and along the way, reimagine queerness vis-à-vis computation.\footnote{Code to support this work can be found at: 
\url{https://github.com/kentchang/subversive-characters-chr2024}.}

\begin{acknowledgments}

We thank the anonymous reviewers for their helpful comments. 
We are grateful for feedback provided by Richard So and Mackenzie Cramer at different stages of this project.
The research reported in this article was supported by funding from the National Science Foundation (IIS-1942591), the National Endowment for the Humanities (HAA-271654-20), and a UC Berkeley Paul Fasana LGBTQ Studies Fellowship to K. C.

\end{acknowledgments}

\bibliography{bibliography,researchstatements}

\begin{thebibliography}{45}
\expandafter\ifx\csname natexlab\endcsname\relax\def\natexlab#1{#1}\fi
\providecommand{\url}[1]{\texttt{#1}}
\providecommand{\href}[2]{#2}
\providecommand{\path}[1]{#1}
\providecommand{\DOIprefix}{doi:}
\providecommand{\ArXivprefix}{arXiv:}
\providecommand{\URLprefix}{URL: }
\providecommand{\Pubmedprefix}{pmid:}
\providecommand{\doi}[1]{\href{http://dx.doi.org/#1}{\path{#1}}}
\providecommand{\Pubmed}[1]{\href{pmid:#1}{\path{#1}}}
\providecommand{\bibinfo}[2]{#2}
\ifx\xfnm\relax \def\xfnm[#1]{\unskip,\space#1}\fi
\bibitem[{Clum(2001)}]{Clum2001-cc}
\bibinfo{author}{J.~M. Clum}, \bibinfo{title}{Something for the Boys: Musical Theater and Gay Culture}, \bibinfo{publisher}{St. Martin's Press}, \bibinfo{year}{2001}.
\bibitem[{Culpeper(2001)}]{Culpeper2001-go}
\bibinfo{author}{J.~Culpeper}, \bibinfo{title}{Language and Characterisation: People in Plays and Other Texts}, \bibinfo{publisher}{Longman}, \bibinfo{year}{2001}.
\bibitem[{Silverstein(2022)}]{Silverstein2022-xc}
\bibinfo{author}{M.~Silverstein}, \bibinfo{title}{Language in Culture: Lectures on the Social Semiotics of Language}, \bibinfo{publisher}{Cambridge University Press}, \bibinfo{year}{2022}.
\bibitem[{Yu et~al.(2020)Yu, Sun, Cardie, and Yu}]{Yu2020-sv}
\bibinfo{author}{D.~Yu}, \bibinfo{author}{K.~Sun}, \bibinfo{author}{C.~Cardie}, \bibinfo{author}{D.~Yu},
\newblock \bibinfo{title}{{Dialogue-Based} {R}elation {E}xtraction},
\newblock in: \bibinfo{editor}{D.~Jurafsky}, \bibinfo{editor}{J.~Chai}, \bibinfo{editor}{N.~Schluter}, \bibinfo{editor}{J.~Tetreault} (Eds.), \bibinfo{booktitle}{Proceedings of the 58th Annual Meeting of the Association for Computational Linguistics}, \bibinfo{publisher}{Association for Computational Linguistics}, \bibinfo{address}{Online}, \bibinfo{year}{2020}, pp. \bibinfo{pages}{4927--4940}.
\bibitem[{Jiang et~al.(2022)Jiang, Xu, Zhan, He, Wang, Xi, Wang, Li, Li, and Yu}]{Jiang2022-bk}
\bibinfo{author}{Y.~Jiang}, \bibinfo{author}{Y.~Xu}, \bibinfo{author}{Y.~Zhan}, \bibinfo{author}{W.~He}, \bibinfo{author}{Y.~Wang}, \bibinfo{author}{Z.~Xi}, \bibinfo{author}{M.~Wang}, \bibinfo{author}{X.~Li}, \bibinfo{author}{Y.~Li}, \bibinfo{author}{Y.~Yu},
\newblock \bibinfo{title}{The {CRECIL} {Corpus}: A {New} {Dataset} for {Extraction} of {Relations} between {Characters} in {Chinese} {Multi-Party} {Dialogues}},
\newblock in: \bibinfo{editor}{N.~Calzolari}, \bibinfo{editor}{F.~B{\'e}chet}, \bibinfo{editor}{P.~Blache}, \bibinfo{editor}{K.~Choukri}, \bibinfo{editor}{C.~Cieri}, \bibinfo{editor}{T.~Declerck}, \bibinfo{editor}{S.~Goggi}, \bibinfo{editor}{H.~Isahara}, \bibinfo{editor}{B.~Maegaard}, \bibinfo{editor}{J.~Mariani}, \bibinfo{editor}{H.~Mazo}, \bibinfo{editor}{J.~Odijk}, \bibinfo{editor}{S.~Piperidis} (Eds.), \bibinfo{booktitle}{Proceedings of the Thirteenth Language Resources and Evaluation Conference}, \bibinfo{publisher}{European Language Resources Association}, \bibinfo{address}{Marseille, France}, \bibinfo{year}{2022}, pp. \bibinfo{pages}{2337--2344}.
\bibitem[{Chen et~al.(2022)Chen, Chu, Wiseman, and Gimpel}]{Chen2022-fg}
\bibinfo{author}{M.~Chen}, \bibinfo{author}{Z.~Chu}, \bibinfo{author}{S.~Wiseman}, \bibinfo{author}{K.~Gimpel},
\newblock \bibinfo{title}{{{{S}umm{S}creen}: A Dataset for Abstractive Screenplay Summarization}},
\newblock in: \bibinfo{booktitle}{{Proceedings of the 60th Annual Meeting of the Association for Computational Linguistics (Volume 1: Long Papers)}}, \bibinfo{publisher}{Association for Computational Linguistics}, \bibinfo{address}{Dublin, Ireland}, \bibinfo{year}{2022}, pp. \bibinfo{pages}{8602--8615}.
\bibitem[{Beltagy et~al.(2020)Beltagy, Peters, and Cohan}]{Beltagy2020-vx}
\bibinfo{author}{I.~Beltagy}, \bibinfo{author}{M.~E. Peters}, \bibinfo{author}{A.~Cohan},
\newblock \bibinfo{title}{{L}ongformer: {T}he {Long-Document} {T}ransformer}  (\bibinfo{year}{2020}). \href{http://arxiv.org/abs/2004.05150}{{\tt arXiv:2004.05150}}.
\bibitem[{Sang et~al.(2022)Sang, Mou, Yu, Yao, Li, and Stanton}]{Sang2022-fv}
\bibinfo{author}{Y.~Sang}, \bibinfo{author}{X.~Mou}, \bibinfo{author}{M.~Yu}, \bibinfo{author}{S.~Yao}, \bibinfo{author}{J.~Li}, \bibinfo{author}{J.~Stanton},
\newblock \bibinfo{title}{{{TVS}how{G}uess}: {C}haracter {C}omprehension in {S}tories as {S}peaker {G}uessing},
\newblock in: \bibinfo{editor}{M.~Carpuat}, \bibinfo{editor}{M.-C. de~Marneffe}, \bibinfo{editor}{I.~V. Meza~Ruiz} (Eds.), \bibinfo{booktitle}{Proceedings of the 2022 Conference of the North American Chapter of the Association for Computational Linguistics: Human Language Technologies}, \bibinfo{publisher}{Association for Computational Linguistics}, \bibinfo{address}{Seattle, United States}, \bibinfo{year}{2022}, pp. \bibinfo{pages}{4267--4287}.
\bibitem[{Wei et~al.(2022)Wei, Wang, Schuurmans, Bosma, Ichter, Xia, Chi, Le, and Zhou}]{Wei2022-io}
\bibinfo{author}{J.~Wei}, \bibinfo{author}{X.~Wang}, \bibinfo{author}{D.~Schuurmans}, \bibinfo{author}{M.~Bosma}, \bibinfo{author}{B.~Ichter}, \bibinfo{author}{F.~Xia}, \bibinfo{author}{E.~Chi}, \bibinfo{author}{Q.~V. Le}, \bibinfo{author}{D.~Zhou},
\newblock \bibinfo{title}{{Chain-of-Thought Prompting Elicits Reasoning in Large Language Models}},
\newblock in: \bibinfo{editor}{S.~Koyejo}, \bibinfo{editor}{S.~Mohamed}, \bibinfo{editor}{A.~Agarwal}, \bibinfo{editor}{D.~Belgrave}, \bibinfo{editor}{K.~Cho}, \bibinfo{editor}{A.~Oh} (Eds.), \bibinfo{booktitle}{{Advances in Neural Information Processing Systems}}, volume~\bibinfo{volume}{35}, \bibinfo{publisher}{Curran Associates, Inc.}, \bibinfo{year}{2022}, pp. \bibinfo{pages}{24824--24837}.
\bibitem[{Cohen(1960)}]{Cohen1960-sb}
\bibinfo{author}{J.~Cohen},
\newblock \bibinfo{title}{A {C}oefficient of {A}greement for {N}ominal {S}cales},
\newblock \bibinfo{journal}{Educational and psychological measurement} \bibinfo{volume}{20} (\bibinfo{year}{1960}) \bibinfo{pages}{37--46}.
\bibitem[{Monroe et~al.(2017)Monroe, Colaresi, and Quinn}]{Monroe2017-us}
\bibinfo{author}{B.~L. Monroe}, \bibinfo{author}{M.~P. Colaresi}, \bibinfo{author}{K.~M. Quinn},
\newblock \bibinfo{title}{{Fightin’} {Words}: {Lexical} {Feature} {Selection} and {Evaluation} for {Identifying} the {Content} of {Political} {Conflict}},
\newblock \bibinfo{journal}{Political analysis: an annual publication of the Methodology Section of the American Political Science Association} \bibinfo{volume}{16} (\bibinfo{year}{2017}) \bibinfo{pages}{372--403}.
\bibitem[{Du et~al.(2019)Du, Liu, and Hu}]{Du2019-ac}
\bibinfo{author}{M.~Du}, \bibinfo{author}{N.~Liu}, \bibinfo{author}{X.~Hu},
\newblock \bibinfo{title}{{Techniques} for {Interpretable} {Machine} {Learning}},
\newblock \bibinfo{journal}{Communications of the ACM} \bibinfo{volume}{63} (\bibinfo{year}{2019}) \bibinfo{pages}{68--77}.
\bibitem[{Sedgwick(1993)}]{Sedgwick1993-bf}
\bibinfo{author}{E.~K. Sedgwick}, \bibinfo{title}{Tendencies}, \bibinfo{publisher}{Duke University Press}, \bibinfo{year}{1993}.
\bibitem[{Halperin(2012)}]{Halperin2012-sa}
\bibinfo{author}{D.~M. Halperin}, \bibinfo{title}{How To Be Gay}, \bibinfo{publisher}{Harvard University Press}, \bibinfo{year}{2012}.
\bibitem[{Pugh(2018)}]{Pugh2018-mo}
\bibinfo{author}{T.~Pugh}, \bibinfo{title}{The Queer Fantasies of the American Family Sitcom}, \bibinfo{publisher}{Rutgers University Press}, \bibinfo{year}{2018}.
\bibitem[{Love(2021)}]{Love2021-jd}
\bibinfo{author}{H.~Love}, \bibinfo{title}{Underdogs}, \bibinfo{publisher}{University of Chicago Press}, \bibinfo{address}{Chicago, IL}, \bibinfo{year}{2021}.
\bibitem[{Jack~Halberstam and Halberstam(2005)}]{Jack_Halberstam2005-jl}
\bibinfo{author}{J.~Jack~Halberstam}, \bibinfo{author}{J.~Halberstam}, \bibinfo{title}{In a Queer Time and Place: Transgender Bodies, Subcultural Lives}, \bibinfo{publisher}{NYU Press}, \bibinfo{year}{2005}.
\bibitem[{Halperin(2019)}]{Halperin2019-hn}
\bibinfo{author}{D.~M. Halperin},
\newblock \bibinfo{title}{Queer love},
\newblock \bibinfo{journal}{Critical inquiry} \bibinfo{volume}{45} (\bibinfo{year}{2019}) \bibinfo{pages}{396--419}.
\bibitem[{Foucault(1998)}]{Foucault1998-um}
\bibinfo{author}{M.~Foucault},
\newblock \bibinfo{title}{Friendship as a {{Way}} of {{Life}}},
\newblock in: \bibinfo{editor}{P.~Rainbow} (Ed.), \bibinfo{booktitle}{Ethics: {{Subjectivity}} and {{Truth}} ({{Essential Works}} of {{Foucault}}, 1954--1984, {{Vol}}. 1)}, \bibinfo{publisher}{NY: The New Press}, \bibinfo{year}{1998}, pp. \bibinfo{pages}{135--140}.
\bibitem[{Davidson(2011)}]{Davidson2011-cw}
\bibinfo{author}{A.~I. Davidson},
\newblock \bibinfo{title}{In {{Praise}} of {{Counter}}-{{Conduct}}},
\newblock \bibinfo{journal}{History of the human sciences} \bibinfo{volume}{24} (\bibinfo{year}{2011}) \bibinfo{pages}{25--41}.
\bibitem[{Raymond(2003)}]{Raymond2003-yi}
\bibinfo{author}{D.~Raymond},
\newblock \bibinfo{title}{{Popular} {Culture} and {Queer} {Representation}},
\newblock \bibinfo{journal}{A Critical Perspective}  (\bibinfo{year}{2003}) \bibinfo{pages}{98--110}.
\bibitem[{McClanahan(2015)}]{McClanahan2015-mj}
\bibinfo{author}{A.~McClanahan},
\newblock \bibinfo{title}{{Disciplining} {Heterosexuality}: {Interrogating} the {Heterosexual} {Ideal}},
\newblock in: \bibinfo{editor}{N.~Farghaly}, \bibinfo{editor}{E.~Leone} (Eds.), \bibinfo{booktitle}{The Sexy Science of The Big Bang Theory: Essays on Gender in the Series}, \bibinfo{publisher}{McFarland}, \bibinfo{year}{2015}, pp. \bibinfo{pages}{88--110}.
\bibitem[{Radloff(2022)}]{Radloff2022-cm}
\bibinfo{author}{J.~Radloff}, \bibinfo{title}{The Big Bang Theory: The Definitive, Inside Story of the Epic Hit Series}, \bibinfo{publisher}{Grand Central Publishing}, \bibinfo{address}{London, England}, \bibinfo{year}{2022}.
\bibitem[{Dumaraog(2021)}]{Dumaraog2021-dk}
\bibinfo{author}{A.~Dumaraog}, \bibinfo{title}{{Big} {Bang} {Theory}: {Kaley} {Cuoco} {Explains} {How} {Penny} {Became} {Less} {Sexualized}}, \bibinfo{howpublished}{\url{https://screenrant.com/big-bang-theory-penny-sexualized-kaley-cuoco-response/}}, \bibinfo{year}{2021}. \bibinfo{note}{Accessed: 2024-10-16}.
\bibitem[{Shaw(2015)}]{Shaw2015-mj}
\bibinfo{author}{J.~Shaw},
\newblock \bibinfo{title}{The adolescent quest},
\newblock in: \bibinfo{editor}{N.~Farghaly}, \bibinfo{editor}{E.~Leone} (Eds.), \bibinfo{booktitle}{The Sexy Science of The Big Bang Theory: Essays on Gender in the Series}, \bibinfo{publisher}{McFarland}, \bibinfo{year}{2015}, pp. \bibinfo{pages}{72--87}.
\bibitem[{Klein(1975)}]{Klein1975-jy}
\bibinfo{author}{M.~Klein}, \bibinfo{title}{Love, Guilt, and Reparation \& Other Works, 1921--1945}, Her the Writings of Melanie Klein, \bibinfo{publisher}{Delacorte Press}, \bibinfo{address}{New York, NY}, \bibinfo{year}{1975}.
\bibitem[{Sedgwick(2003)}]{Sedgwick2003-zk}
\bibinfo{author}{E.~K. Sedgwick}, \bibinfo{title}{Touching {{Feeling}}}, \bibinfo{publisher}{Duke University Press}, \bibinfo{year}{2003}.
\bibitem[{Bradway(2021)}]{Bradway2021-he}
\bibinfo{author}{T.~Bradway},
\newblock \bibinfo{title}{{Queer} {Narrative} {Theory} and the {Relationality} of {Form}},
\newblock \bibinfo{journal}{Publications of the Modern Language Association of America} \bibinfo{volume}{136} (\bibinfo{year}{2021}) \bibinfo{pages}{711--727}.
\bibitem[{Halberstam(2011)}]{Halberstam2011-lf}
\bibinfo{author}{J.~Halberstam}, \bibinfo{title}{The Queer Art of Failure}, \bibinfo{publisher}{Duke University Press}, \bibinfo{year}{2011}.
\bibitem[{Warner(1993)}]{Warner1993-qw}
\bibinfo{author}{M.~Warner}, \bibinfo{title}{Fear of a Queer Planet}, Studies in Classical Philology, \bibinfo{publisher}{University of Minnesota Press}, \bibinfo{address}{Minneapolis, MN}, \bibinfo{year}{1993}.
\bibitem[{Marcus et~al.(2016)Marcus, Love, and Best}]{Marcus2016-hc}
\bibinfo{author}{S.~Marcus}, \bibinfo{author}{H.~Love}, \bibinfo{author}{S.~Best},
\newblock \bibinfo{title}{{Building} a {Better} {Description}},
\newblock \bibinfo{journal}{Representations} \bibinfo{volume}{135} (\bibinfo{year}{2016}) \bibinfo{pages}{1--21}.
\bibitem[{Britzman(2013)}]{Britzman2013-wa}
\bibinfo{author}{D.~P. Britzman},
\newblock \bibinfo{title}{Theory {K}indergarten},
\newblock in: \bibinfo{editor}{S.~M. Barber}, \bibinfo{editor}{D.~L. Clark} (Eds.), \bibinfo{booktitle}{Regarding Sedgwick}, \bibinfo{publisher}{Routledge}, \bibinfo{address}{London, England}, \bibinfo{year}{2013}, pp. \bibinfo{pages}{121--142}.
\bibitem[{Love(2010)}]{Love2010-br}
\bibinfo{author}{H.~Love},
\newblock \bibinfo{title}{{Truth} and {Consequences}: {On} {Paranoid} {Reading} and {Reparative} {Reading}},
\newblock \bibinfo{journal}{Criticism} \bibinfo{volume}{52} (\bibinfo{year}{2010}) \bibinfo{pages}{235--241}.
\bibitem[{Chang(2023)}]{Chang2023-ff}
\bibinfo{author}{K.~K. Chang},
\newblock \bibinfo{title}{{The} {Queer} {Gap} in {Cultural} {Analytics}},
\newblock in: \bibinfo{editor}{M.~K.~G. Lauren F.~Klein} (Ed.), \bibinfo{booktitle}{Debates in the Digital Humanities 2023}, \bibinfo{publisher}{U of Minnesota Press}, \bibinfo{year}{2023}, pp. \bibinfo{pages}{105--119}.
\bibitem[{Silverstein(2004)}]{Silverstein2004-em}
\bibinfo{author}{M.~Silverstein},
\newblock \bibinfo{title}{{``Cultural''} {C}oncepts and the {Language‐Culture} {N}exus},
\newblock \bibinfo{journal}{Current anthropology} \bibinfo{volume}{45} (\bibinfo{year}{2004}) \bibinfo{pages}{621--652}.
\bibitem[{Li et~al.(2024)Li, Xu, Shang, Liu, Ji, and Guo}]{Li2024-db}
\bibinfo{author}{G.~Li}, \bibinfo{author}{Z.~Xu}, \bibinfo{author}{Z.~Shang}, \bibinfo{author}{J.~Liu}, \bibinfo{author}{K.~Ji}, \bibinfo{author}{Y.~Guo},
\newblock \bibinfo{title}{{Empirical} {Analysis} of {Dialogue} {Relation} {Extraction} with {Large} {Language} {Models}}  (\bibinfo{year}{2024}). \href{http://arxiv.org/abs/2404.17802}{{\tt arXiv:2404.17802}}.
\bibitem[{Sun et~al.(2017)Sun, Schiele, and Fritz}]{Sun2017-al}
\bibinfo{author}{Q.~Sun}, \bibinfo{author}{B.~Schiele}, \bibinfo{author}{M.~Fritz},
\newblock \bibinfo{title}{{A} {Domain} {Based} {Approach} to {Social} {Relation} {Recognition}},
\newblock in: \bibinfo{booktitle}{2017 {IEEE} Conference on Computer Vision and Pattern Recognition ({CVPR})}, \bibinfo{publisher}{IEEE}, \bibinfo{year}{2017}.
\bibitem[{Lu et~al.(2022)Lu, Hu, Cheng, Smith, and Ostendorf}]{Lu2022-ws}
\bibinfo{author}{B.-R. Lu}, \bibinfo{author}{Y.~Hu}, \bibinfo{author}{H.~Cheng}, \bibinfo{author}{N.~A. Smith}, \bibinfo{author}{M.~Ostendorf},
\newblock \bibinfo{title}{{Unsupervised} {Learning} of {Hierarchical} {Conversation} {Structure}},
\newblock in: \bibinfo{editor}{Y.~Goldberg}, \bibinfo{editor}{Z.~Kozareva}, \bibinfo{editor}{Y.~Zhang} (Eds.), \bibinfo{booktitle}{Findings of the Association for Computational Linguistics: {EMNLP} 2022}, \bibinfo{publisher}{Association for Computational Linguistics}, \bibinfo{address}{Abu Dhabi, United Arab Emirates}, \bibinfo{year}{2022}, pp. \bibinfo{pages}{5657--5670}.
\bibitem[{Liu et~al.(2023)Liu, Zhang, Zhang, Xue, and You}]{Liu2023-lp}
\bibinfo{author}{X.~Liu}, \bibinfo{author}{J.~Zhang}, \bibinfo{author}{H.~Zhang}, \bibinfo{author}{F.~Xue}, \bibinfo{author}{Y.~You},
\newblock \bibinfo{title}{{Hierarchical} {Dialogue} {Understanding} with {Special} {Tokens} and {Turn-Level} {Attention}},
\newblock \bibinfo{journal}{Tiny Papers @ ICLR}  (\bibinfo{year}{2023}).
\bibitem[{Stache and Davidson(2019)}]{Stache2019-ea}
\bibinfo{author}{L.~C. Stache}, \bibinfo{author}{R.~D. Davidson}, \bibinfo{title}{Gilmore Girls: A Cultural History}, \bibinfo{publisher}{Rowman \& Littlefield}, \bibinfo{year}{2019}.
\bibitem[{Richardson(2010)}]{Richardson2010-kg}
\bibinfo{author}{K.~Richardson}, \bibinfo{title}{{Television Dramatic Dialogue: A Sociolinguistic Study}}, \bibinfo{publisher}{Oxford University Press}, \bibinfo{year}{2010}.
\bibitem[{Bednarek(2023)}]{Bednarek2023-zu}
\bibinfo{author}{M.~Bednarek}, \bibinfo{title}{Language and Characterisation in Television Series: A Corpus-informed Approach to the Construction of Social Identity in the Media}, \bibinfo{publisher}{John Benjamins Publishing Company}, \bibinfo{year}{2023}.
\bibitem[{Kozloff(2000)}]{Kozloff2000-pe}
\bibinfo{author}{S.~Kozloff}, \bibinfo{title}{{Overhearing Film Dialogue}}, \bibinfo{publisher}{University of California Press}, \bibinfo{year}{2000}.
\bibitem[{Butler(1990)}]{Butler1990-rb}
\bibinfo{author}{J.~Butler}, \bibinfo{title}{Gender Trouble: Feminism and the Subversion of Identity}, \bibinfo{publisher}{Routledge}, \bibinfo{address}{NY}, \bibinfo{year}{1990}.
\bibitem[{Love(2015)}]{Love2015-jo}
\bibinfo{author}{H.~Love},
\newblock \bibinfo{title}{{Doing} {Being} {Deviant}: {Deviance} {Studies}, {Description}, and the {Queer} {Ordinary}},
\newblock \bibinfo{journal}{Differences} \bibinfo{volume}{26} (\bibinfo{year}{2015}) \bibinfo{pages}{74--95}.

\end{thebibliography}

\appendix

\newpage

\section{Additional related work}\label{sec:related_work}

This work builds on prior research in several disciplinary traditions.

\paragraph{Sociolinguistics and linguistic anthropology.} The study of social relationships through language is central to the fields of sociolinguistics and linguistic anthropology. 
These disciplines provide tools for analyzing how language constructs and reflects social identities and power dynamics.
For this work, Michael Silverstein's work~\cite{Silverstein2004-em,Silverstein2022-xc}, which explores the ways in which language ideologies and linguistic practices intersect, informs our approach to how social relationships are subverted and maintained through dialogue.

\paragraph{Dialogue understanding and natural language processing.} The design and implementation of our computational methods are indebted to the field of natural language processing (NLP), especially work on dialogue understanding.
Our main task is built on dialogue relation extraction techniques, which are employed to classify the relationships between characters based on their conversational exchanges~\cite{Jiang2022-bk,Li2024-db}. 
This work also builds upon narrative understanding and conversation modeling.
This is exemplified by, among others, TVShowGuess~\cite{Sang2022-fv}, which leverages neural models to perform reading comprehension tasks on television shows. 
In the context of this work, understanding the hierarchical structure of dialogue is also crucial%
~\cite{Sun2017-al,Lu2022-ws,Liu2023-lp}. 

\paragraph{TV and film studies.} Much of this work aims to understand representation on screen, which is the subject of TV and film studies~\cite{Stache2019-ea}, and in particular, analyzing dialogue within TV and film is essential for understanding how social relationships are portrayed and subverted. 
This work is inspired by linguistic analysis in this space~\cite{Richardson2010-kg,Bednarek2023-zu} and works such as Sarah Kozloff's~\cite{Kozloff2000-pe} which look into the mechanics of scripted dialogue and its impact on the audience's perception of characters and their relationships.

\paragraph{Gender and queer studies.} The subversion of social relationships in TV often intersects with issues of gender and sexuality. 
Judith Butler's theories on gender performativity provide the essential framework for understanding how characters on TV subvert traditional gender roles and expectations~\cite{Butler1990-rb}. 
The positivist approaches in queer studies~\cite{Halperin2012-sa,Love2015-jo} inform the theoretical foundation of this work as we articulate subversion in computational terms.

\section{Ranked relationship types}\label{sec:rel_types}

See Table~\ref{tab:relationship_types}.

\begin{table}[!ht]
    \centering
    \caption{Ranked relationship types.}
    \label{tab:relationship_types}    
    \begin{tabular}{@{}rl@{\hskip 10pt}|@{\hskip 10pt}rl@{}}
        \toprule
        \textbf{Rank} & \textbf{Relationship Type} & \textbf{Rank} & \textbf{Relationship Type} \\
        \midrule
        1 & grandparent\_of & 15 & enemy\_of \\
        2 & grandchild\_of & 16 & colleague\_of \\
        3 & parent\_of & 17 & classmate\_of \\
        4 & child-in-law\_of & 18 & roommate\_of \\
        5 & child\_of & 19 & neighbor\_of \\
        6 & sibling-in-law\_of & 20 & teacher\_of \\
        7 & sibling\_of & 21 & student\_of \\
        8 & relative\_of & 22 & boss\_of \\
        9 & ex-spouse\_of & 23 & subordinate\_of \\
        10 & ex-boy/girlfriend\_of & 24 & trainer\_of \\
        11 & ex-love\_interest\_of & 25 & trainee\_of \\
        12 & spouse\_of & 26 & acquaintance\_of \\
        13 & boy/girlfriend\_of & 27 & friend\_of \\
        14 & love\_interest\_of & 28 & other \\
        \bottomrule
    \end{tabular}
\end{table}

\section{Dataset statistics}\label{sec:data}

After the process described in Section~\ref{sec:methods}, each pilot teleplay is now transformed into a sequence of scenes, which is comprised of speakers with canonical names and their lines, as well as, where applicable, a list of relationship tuples for speakers in the scene.
The statistics of title and token counts for this dataset are reported in Table~\ref{tab:summary}.\footnote{Tokens are counted with the tiktoken implementation of BPE tokenizer \texttt{o200k\_base}:~\url{https://github.com/openai/tiktoken/tree/main}.}

\begin{table}[!ht]
    \centering
    \caption{Summary of train, development, and test sets.}
    \label{tab:summary}    
\begin{tabular}{@{}lrrr@{}}
\toprule
                               & \textbf{training} & \textbf{development} & \textbf{test} \\ \midrule
\# titles                      & 552               & 115                  & 120           \\
\# scenes                      & 36,320            & 7,078                & 6,965         \\
\# number of dyads with labels & 9,223             & 4,978                & 7,176         \\
\# avg tokens per              &                   &                      &               \\
\;\;\;scene                          & 192               & 194                  & 201           \\
\;\;\;utterance                      & 95                & 96                   & 103           \\ \bottomrule
\end{tabular}
\end{table}

\section{Sample prompts}\label{sec:llamaprompt}

\subsection{LLaMA 3 prompt}

See Fig.~\ref{fig:prompt}.

\begin{figure}[h!]
    \centering
\begin{lstlisting}[frame=single, basicstyle=\ttfamily\footnotesize, breaklines=true, escapeinside={(*@}{@*)}]
messages = [
    {
        "role": "system",
        "content": f"""Your goal is to extract relationships between TV characters in a scene of a TV series.
You will be provided with their dialogues, wrapped in <dialogue>.
Speaker names start with `ENTITY`, and their lines are separated by `:`.
You will read the dialogue and identify the relationship between a certain pair of entities, as requested in <question>.
The relationship is directed, so the order of entities in each triplet matters.
Here are the possible relationship types: {LABEL_OPTIONS}.
Here is an example:"""
    },
    {
        "role": "user",
        "content": f"""<dialogue>SCENE: INT. WEINBERG APARTMENT - MIDGE'S OLD BEDROOM - MOMENTS LATER

ENTITY 24: That forehead is not improving.

[ENTITY 24 lifts ESTHER out and lays her down on the bed.]

ENTITY 2: What? Are you sure?
ENTITY 24: It's getting bigger. The whole face will be out of proportion.
ENTITY 2: But look at her nose. It's elongating now, see?
ENTITY 24: The nose is not the problem. The nose you can fix. But this gigantic forehead...
ENTITY 2: Well, there's always bangs.
ENTITY 24: I'm just afraid she's not a very pretty girl.
ENTITY 2: Mama, she's a baby.
ENTITY 24: I just want her to be happy. It's easier to be happy when you're pretty.
ENTITY 24: You're right. Bangs will help.</dialogue>
<question> ENTITY 2 is what of ENTITY 24? ANSWER with ONLY {LABEL_OPTIONS} </question>"""
    },
    {
        "role": "assistant",
        "content": "child_of"
    },
    {
        "role": "system",
        "content": f"""Great job! You have successfully identified the relationship between the two entities. Now, let's move on to the next one."""
    },
    {
        "role": "user",
        "content": f"""<dialogue>{scene_string}</dialogue>
<question> {head} is what of {tail}? ANSWER with ONLY: {LABEL_OPTIONS}.</question>"""
    },
]
\end{lstlisting}
    \caption{Ones-shot prompt for LLaMA 3--70b--instruct.}\label{fig:prompt}
\end{figure}

\subsection{Sample OpenAI o1-mini prompt and response}

\begin{figure}%
  \centering
  \includegraphics[width=0.92\linewidth]{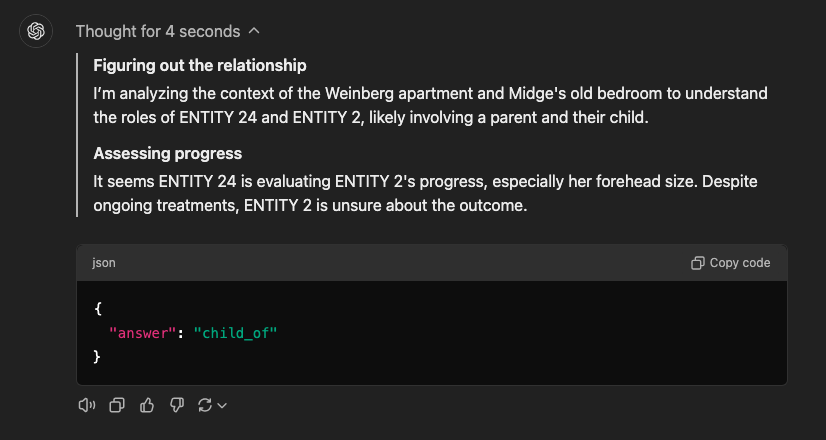} %
  \caption{An example of prompting ChatGPT o1-mini.}\label{fig:o1response}
\end{figure}

See Fig.~\ref{fig:prompt2} for the prompt, and Fig.~\ref{fig:o1response} for an example response from ChatGPT o1-mini. 
The API to o1-mini does not include access to the tokens used for chain-of-thought, so this example is included as a sanity check and see the thinking process of o1 models where the thought process is visible to us.

\begin{figure}[h!]
    \centering
\begin{lstlisting}[frame=single, basicstyle=\ttfamily\footnotesize, breaklines=true, escapeinside={(*@}{@*)}]
messages = [
    {
        "role": "system",
        "content": f"""You are a helpful assistant designed to extract relationships between TV characters in a scene of a TV series.
You will be provided with their dialogues, wrapped in <dialogue>.
Speaker names start with `ENTITY`, and their lines are separated by `:`.
You will read the dialogue and identify the relationship between a certain pair of entities, as requested in <question>.
The relationship is directed, so the order of entities in each triplet matters.

**Return only a JSON object** with the following property:

- "answer": one of the following {LABEL_OPTIONS}.

This property must always be present. 

Do not include any additional text or explanations outside the JSON object.

<dialogue>SCENE: INT. WEINBERG APARTMENT - MIDGE'S OLD BEDROOM - MOMENTS LATER

ENTITY 24: That forehead is not improving.

[ENTITY 24 lifts ESTHER out and lays her down on the bed.]

ENTITY 2: What? Are you sure?
ENTITY 24: It's getting bigger. The whole face will be out of proportion.
ENTITY 2: But look at her nose. It's elongating now, see?
ENTITY 24: The nose is not the problem. The nose you can fix. But this gigantic forehead...
ENTITY 2: Well, there's always bangs.
ENTITY 24: I'm just afraid she's not a very pretty girl.
ENTITY 2: Mama, she's a baby.
ENTITY 24: I just want her to be happy. It's easier to be happy when you're pretty.
ENTITY 24: You're right. Bangs will help.</dialogue>
<question> ENTITY 2 is what of ENTITY 24?</question>"""
    }
]
\end{lstlisting}
    \caption{Prompt for OpenAI o1-mini.}\label{fig:prompt2}
\end{figure}

\end{document}